\def\year{2021}\relax
\documentclass[letterpaper]{article} 
\usepackage{aaai21}  
\usepackage{times}  
\usepackage{helvet} 
\usepackage{courier}  
\usepackage[hyphens]{url}  
\usepackage{graphicx} 
\usepackage{colortbl}
\usepackage{hhline}
\usepackage{array}
\usepackage{amssymb}
\usepackage{ulem}
\usepackage{natbib}  
\usepackage{caption} 
\usepackage{amsmath,amsfonts,bm}

\usepackage{hyperref}
\usepackage{url}
\usepackage{todonotes}
\presetkeys%
    {todonotes}%
    {inline,backgroundcolor=yellow}{}
\usepackage{subcaption}
\usepackage{booktabs}
\usepackage{multirow}
\usepackage{adjustbox}

\usepackage[english]{babel}
\usepackage{blindtext}

\title{Synthesising Electronic Health Records: Cystic Fibrosis Patient Group}
\author {
    Emily Muller,\textsuperscript{\rm 1}
    Xu Zheng, \textsuperscript{\rm 2}
    Jer Hayes \textsuperscript{\rm 2} \\
}
\affiliations {
    \textsuperscript{\rm 1} Department of Epidemiology and Biostatistics, Imperial College London, UK \\
    \textsuperscript{\rm 2} Accenture Labs, Dublin, Ireland \\
    emuller@ic.ac.uk, 
    xu.b.zheng@accenture.com, 
    jeremiah.hayes@accenture.com
}
\begin{document}

\maketitle

\begin{abstract}
Class imbalance can often degrade predictive performance of supervised learning algorithms. Balanced classes can be obtained by oversampling exact copies, with noise, or interpolation between nearest neighbours (as in traditional SMOTE methods). Oversampling using augmentation, as is typical in computer vision tasks, can be achieved with deep generative models. Deep generative models are effective data synthesisers due to their ability to capture complex underlying distributions. Synthetic data in healthcare can enhance interoperability between healthcare providers by ensuring patient privacy. Equipped with large synthetic datasets which do well to represent small patient groups, machine learning in healthcare can address the current challenges of bias and generalisability. This paper evaluates synthetic data generators ability to synthesise patient electronic health records. We test the utility of synthetic data for patient outcome classification, observing increased predictive performance when augmenting imbalanced datasets with synthetic data.
\end{abstract}

\section{Introduction}\label{intro}
Massive amounts of data are generated in the field of medicine from sources such as biosensors, physiological measurements, genome sequencing and electronic health records (EHR). Machine learning is uniquely placed to address pattern recognition in the medical field, whether helping to interpret medical scans, predicting disease trajectory, improving health system workflows or promoting patient data sovereignty \citep{topol2019high}. Large datasets that are representative and diverse are needed to mitigate bias. However, aggregating data in this way can be; expensive, if the data is proprietary; inaccessible, due to interoperability standards in the sharing of health data; or illegal as it has the potential to violate privacy. To overcome this, synthetic data are increasingly being used in the healthcare setting \citep{chen2021synthetic}. Synthetic data generation of EHRs for underrepresented patient groups or small patient datasets can enable better predictive modelling while preserving patient privacy. In this work, we collect Cystic Fibrosis EHRs and synthesise new patient records. We examine the utility of four synthetic data generators and augment the original data with synthetic data to enhance predictability in patient outcomes.

Synthetic data generators are algorithms used generate new samples that preserve the original data distribution while adhering to numerous properties, namely: utility and privacy. Utility measures the effectiveness of the synthetic data on another task, risk prediction for example. Privacy ensures that the resulting synthetic data cannot be used to infer anything about any single individual. Differential privacy provides an algorithmic formulation of privacy \cite{dwork2014algorithmic}. Since our original data is anonymised EHRs with non-sensitive features, we consider instead of privacy the requirement of \textit{uniqueness}: to not simply copy the input data. 

Data augmentation has shown increased performance for image related deep learning tasks\cite{shorten2019survey}. Traditionally, tabular augmentation methods, such as Synthetic Minority Over-sampling Technique (SMOTE) and its variants, synthesise tabular data as midpoints to k-nearest neighbours \cite{chawla2002smote}. Another simple and effective method of synthetic data generation is through perturbation of the original data by adding noise (winner of Hide-and-seek challenge, 2020 \cite{alaa2021faithful}). Deep learning methods for tabular data synthesis involve generative models such as generative adversarial networks (GANs) and variational autoencoders (VAEs) \cite{xu2018synthesizing, choi2017generating, xu2019modeling, xie2018differentially, yoon2020anonymization}. These models implicitly parameterise the multivariate distribution of the original data using deep neural networks. In \cite{fiore2019using}, authors use a GAN to synthesise the underrepresented class of fraudulent credit card cases, observing a maximum increase in classification sensitivity of $3.4$ percentage points when augmenting the small class 3 times its size ($0.55\%$ of the training set). A conditional GAN (cGAN) has shown optimal performance on a set of 22 tabular datasets \cite{douzas2018effective}. However, conditioning on the outcome class is not always desirable or available in practice. In an application of augmenting classes of thermal comfort, the authors find 2 experiments where synthetic data alone has a higher F1 score than the original training data. Similar to \cite{fiore2019using}, the author synthesises the underrepresented class hence making known the class label. Other work has shown the efficacy of generative networks above traditional methods \cite{liu2019wasserstein, ngwenduna2021alleviating, engelmann2020conditional}. However, we do not find any studies reporting the similarity of the synthesisers to the original dataset. This is especially relevant for deep generative models which can suffer mode collapse and learn to copy the training data, thus, violating privacy assurances. The high capacity of generative models makes them a good candidate for capturing complex non-linear distributions in the original data, while their intractable likelihood functions make evaluation difficult. We employ model-agnostic sample-wise similarity metrics to evaluate model performance at each training epoch. Similarity and uniqueness are calculated for each synthetic dataset and we compare this to utility in a classification task. We find that synthetic augmentation improves accuracy in two out of four models, while there are trade-offs in similarity. In the next section, we detail the feature extraction procedure, model selection, and similarity and uniqueness metrics.

\section{Method}\label{method}
The complex trajectory for Cystic Fibrosis patients requires close monitoring of care and high-risk decision making, such as lung transplant referral. Opportunities for individualised patient care using machine learning have recently been put forward for the case of Cystic Fibrosis (CF) patients \citep{abroshan2020opportunities}. The authors highlight risk prediction and personalised treatment recommendation, amongst others, as areas where data can be leveraged by machine learning. Traditionally, research on risk prediction has used regression models, see review article \cite{breuer2018predicting}. More recently, machine learning method AutoPrognosis was able to outperform 3 regression models and the clinical standard of measured forced expiratory volume $FEV_1\%<30$ by taking a feature-agnostic approach and including a set of 115 patient registry variables \cite{alaa2018autoprognosis}. We explore predictive modeling for CF patients using synthetically augmented EHR records.

\paragraph{Feature extraction. }
Cystic Fibrosis patients are extracted from the IBM Explorys database. A total of $10074$ patients are extracted, representing about 1/3 ($31199$) of all CF patients in the US \citep{USCF}. Patients belong to two subgroups: having died or having received a lung transplant, or having survived. We predict outcome based on a set of features including: demographic, comorbidity, lung infection and therapy variables. We find large discrepancies in the prevalance of comorbidities, lung infections and therapy variables between our patient group and those reported in the UK CF Registry \citep{alaa2018autoprognosis} and the US CF Foundation \cite{USCF} (see Appendix Table \ref{cf_diff}). Since the IBM Explorys database are EHRs based on medical health insurance it is plausible that entire diagnosis codes go unrecorded if they are irrelevant for a medical claim. A lung transplant is an expensive procedure, hence we assume there is no missingness in this outcome class. Missingness of features in this dataset should be treated carefully for clinical research. However, since we are evaluating synthetic data generators, we continue to use this CF patient dataset with its caveats.

\paragraph{Pre-processing. }We remove all demographic or measurement variables with greater than 50\% missingness and uncertain encoding, resulting in a binary matrix of 41 variables (see Appendix Table \ref{cf_diff}). In order to enhance synthetic diversity, we remove all samples with no diagnosis codes and all duplicates. Our final dataset has $3184$ patients, with $\sim 80\%$ belonging to the survived subgroup.

\paragraph{Synthetic Data Generators. } We create 4 synthetic datasets using VAE, Differentially Private GAN (DPGAN) with two noise levels \citep{xie2018differentially} and Conditional GAN (CTGAN) \citep{xu2019modeling}. Our VAE is trained with 2 hidden layers in both the encoder and decoder and ELBO loss minimisation. DPGAN is trained as a typical minimax game between the discriminator and generator where noise is added to the gradient during training. We employ the cross-entropy derived loss. The authors show addition of noise in this way guarantees a level or differential privacy. CTGAN addresses sparsity and imbalanced categorical columns by sampling vectors during training and introducing a condition for which the generator learns the conditional distribution. We optimise each model using model-agnostic similarity metrics (see \nameref{experimental_details}).

\paragraph{Similarity. } Similarity of the synthetic dataset with the original dataset is measured in two ways. Fidelity measures the degree to which generated samples resemble the original data and diversity measures whether generated samples cover the full variability of the original data. The latter is particularly useful for evaluating generative models which are prone to mode collapse. Precision and recall metrics for evaluating generative models were originally proposed in \citep{sajjadi2018assessing}. Precision is defined as the proportion of the synthetic probability distribution that can be generated by the original probability distribution, thus measuring fidelity, and recall symmetrically defines diversity. Precision and recall are effectively calculated as the proportion of samples which lie in the support of the comparative distribution. This assumes uniform density across the support and therefore alternative metrics for precision and recall have been proposed to ameliorate this \citep{kynkaanniemi2019improved, naeem2020reliable, alaa2021faithful}. We employ the original precision and recall metrics along with density and coverage metrics from \citep{naeem2020reliable}. Density and coverage addresses the lack of robustness to outliers, failure to detect matching distributions and inability to diagnose different types of distribution failure. 

\begin{figure*}
     \begin{subfigure}[b]{0.33\textwidth}
         \centering
         \includegraphics[width=\textwidth]{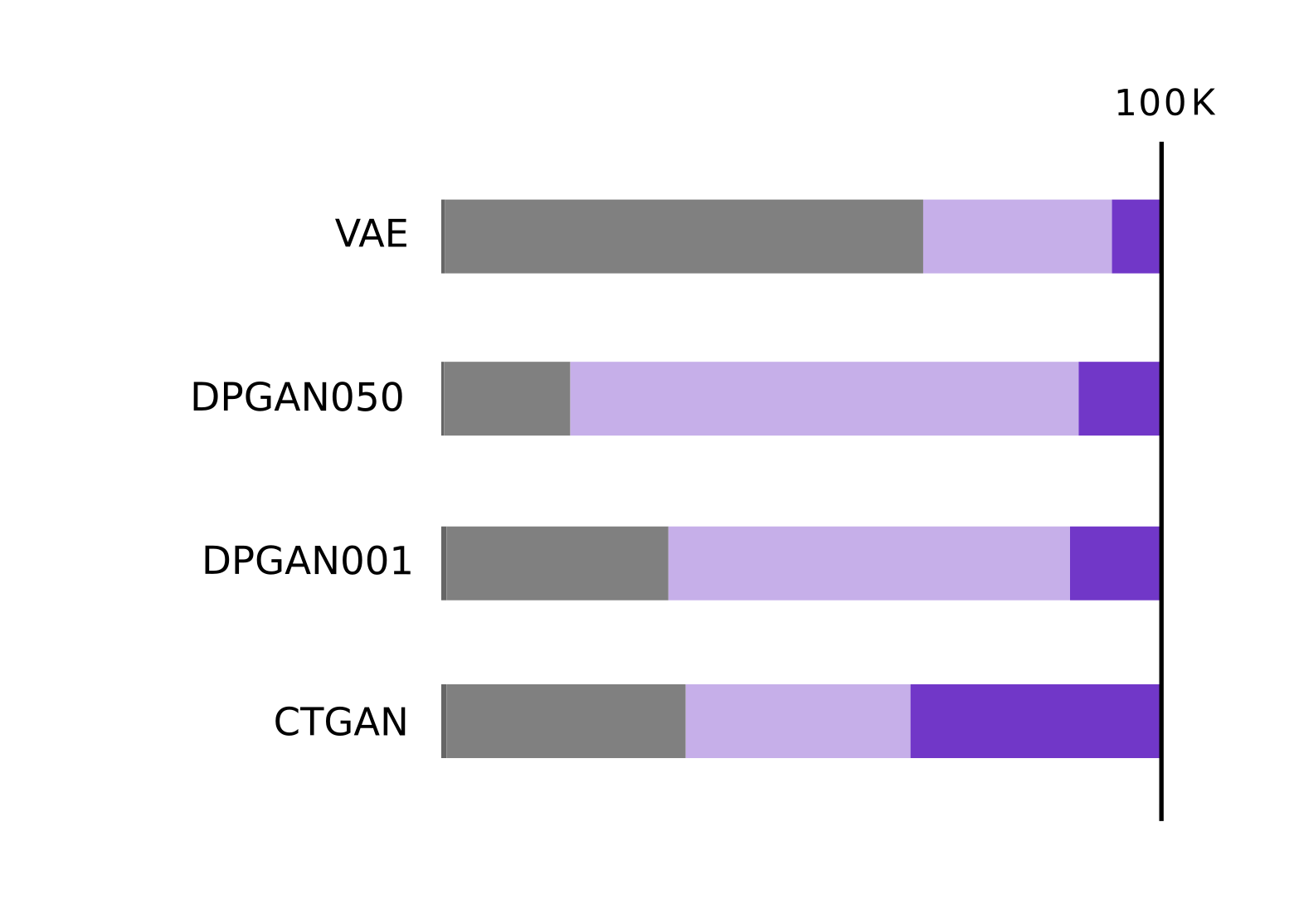}
         \caption{Authenticity proportion}
         \label{fig1:privacy}
     \end{subfigure}
     \hfill
     \begin{subfigure}[b]{0.33\textwidth}
         \centering
         \includegraphics[width=\textwidth]{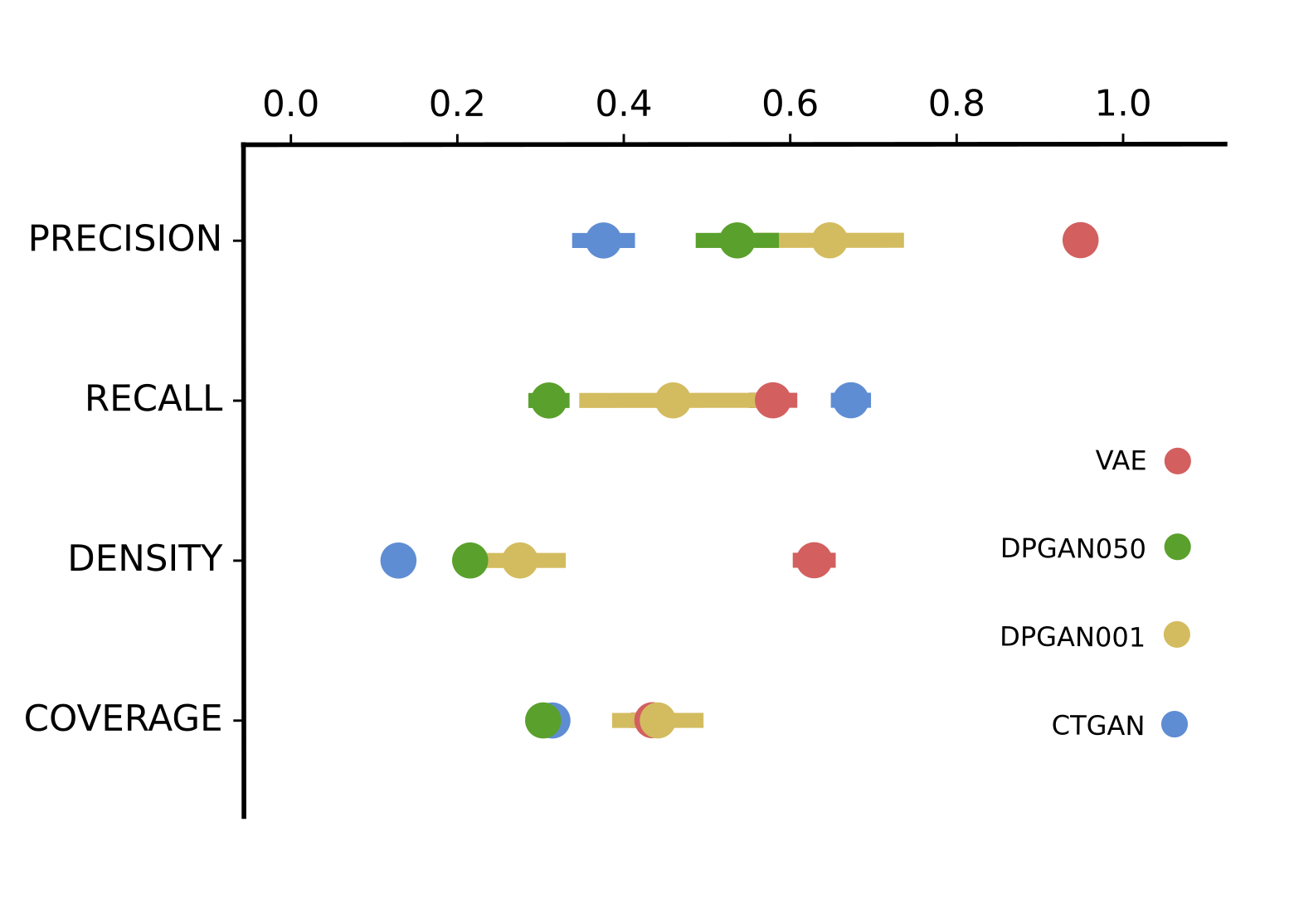} 
         \caption{Similarity metrics}
         \label{fig2:similarity}
     \end{subfigure}
     \begin{subfigure}[b]{0.33\textwidth}
         \centering
         \includegraphics[width=\textwidth]{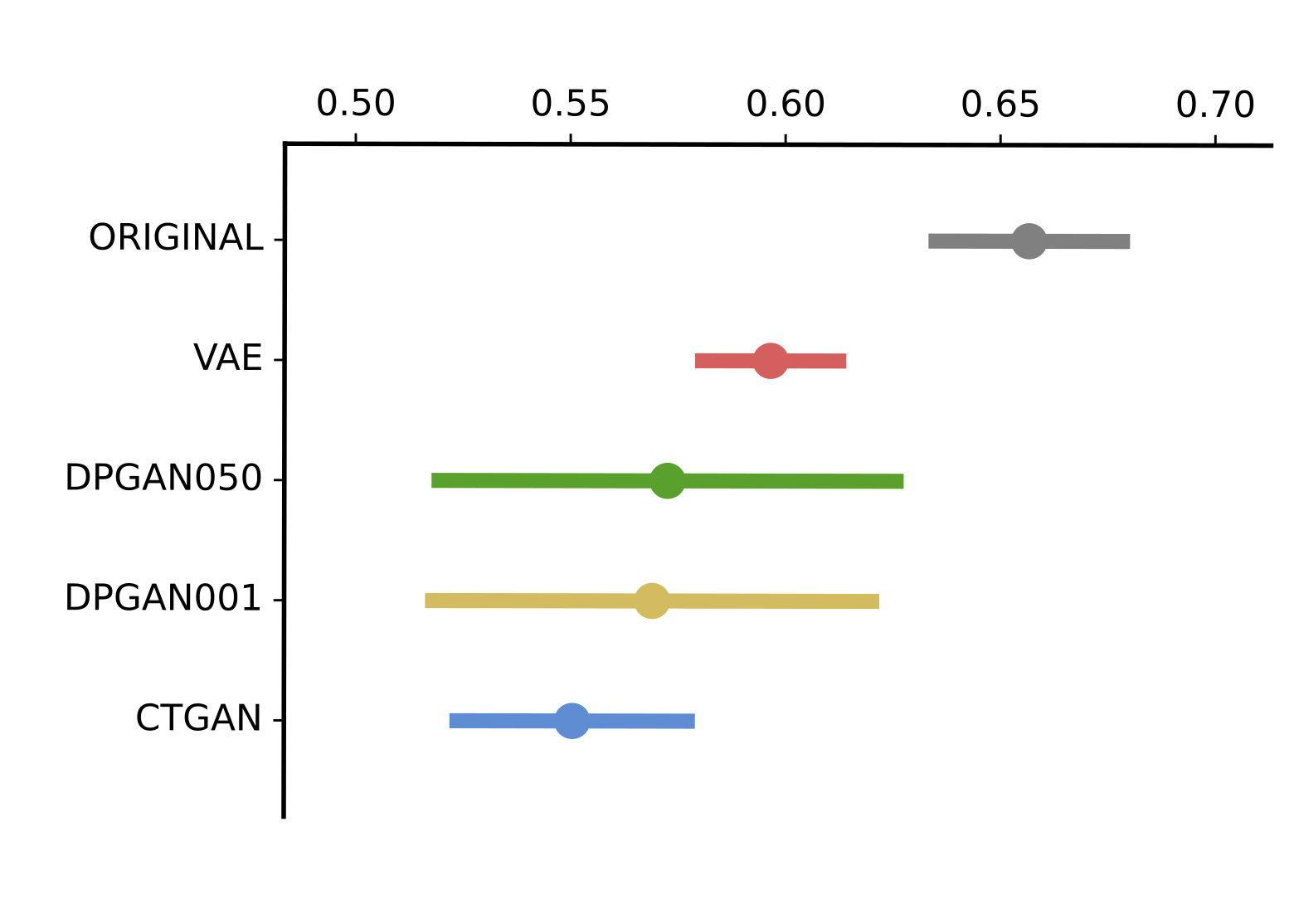}
         \caption{Synthetic data AUC-ROC}
         \label{fig3:synth_auc}
     \end{subfigure}
    \caption{(A) Authenticity of $100k$ samples from each generator. Grey are proportion of samples which appear in the original training data, with darker grey (far left) representing those which are unique. Light purple are samples which do not appear in the original training data, and darker purple represent those which are unique. (B) Similarity metrics for each model. For each fold, a dataset matching the size of the original fold with the equivalent proportion of classes is sampled from the unique synthetic dataset (dark blue only). This is repeated 10 times and similarity metrics show mean and standard deviation over folds and repetitions.}
    \label{fig:stand_alone}
\end{figure*}

\paragraph{Uniqueness. }
A synthetic data generator can achieve perfect fidelity and diversity by copying the original training data. Privacy assurances are essential to prevent leakage of personal information. Differential privacy is one such well known and commonly researched assurance \cite{dwork2014algorithmic}. Differentially private algorithms limit how much; the output can differ based on whether input is included or not; one can learn about a person because their data was included; and confidence about whether someone's data was included. In practice, there are various distance based metrics to gaurantee such assurance. In \cite{alaa2021faithful}, the authors quantify a generated sample as authentic if its closest real sample is at least closer to any other real sample than the generated sample. Extending this to the case of binary variables, we could consider hamming distance, however, since the IBM Explorys data is de-personalised and non-identifiable, we assess our generators based on how many exact copies are made. We calculate the uniqueness of each model by generating a large finite number of samples, and reporting the percentage overlap with the original training data, and the uniqueness of samples. In post-processing, we remove all generated samples which copy the original training data and remove all duplicates. 

\paragraph{Utility.} To empirically validate the utility of the generated dataset we introduce two different training testing settings. \textbf{Setting A:} train the predictive models on the synthetic training set, test the performance of the models on the real testing set. \textbf{Setting B:} train on the synthetically augmented balanced-class training set, test on the real testing set. We performance 5-fold cross validation, sampling each fold with proportional representation of each class. After training the synthesisers (described below), we obtain unique synthetic datasets for each fold and for each model. Setting A was implemented as follows: for each fold, we sampled a synthetic dataset the size of the original dataset with equivalent class proportions to calculate utility. We repeated this 10 times resulting in 50 measurements for each model for which we report the mean and standard deviation. For each synthetic dataset we trained the following classification models: Support Vector Machine, Linear Regression, Naive Bayes, K-Nearest Neighbours and a Random Forest. We chose the classifier which maximised area under the receiver operating curve (AUC-ROC) for the hold out test set and compared this to training with the original data. As well as testing utility on the synthetic dataset alone, we also augmented the original dataset to obtain class balance (Setting B). Not all synthesisers provided enough unique samples of the small class, and we therefore attempted to upsample as far as possible, otherwise downsampling the larger class. With a balanced class augmented training set, we performed the same utility experiment as outlined above reporting both area under the receiver operating curve (AUC-ROC) and accuracy.

\paragraph{Synthesisers. } We optimised each synthesiser on the sum of precision, recall, density and coverage for a random 5th of the dataset stratified by outcome class. After hyperparameter tuning, we trained each model for each of the 5 separate training folds and generated $100K$ synthetic samples from each. We post-processed each synthetic dataset to obtain only unique samples. For each fold, we sampled a synthetic dataset the size of the original dataset with equivalent class proportions to calculate similarity. We repeated this 10 times resulting in 50 measurements for each metric and model for which we report the mean and standard deviation (see Appendix \nameref{experimental_details}).

\paragraph{Implementation. }Synthesisers were implemented on a nvidia-gpu. Utility models were implemented on a CPU using \textit{python} and the \textit{sklearn.classification} module. Similarity metrics were calculuted using the python module\textit{ prdc} \cite{naeem2020reliable} with $k=5$.

\begin{figure*}
     \begin{subfigure}[b]{0.5\textwidth}
         \raggedleft
         \includegraphics[width=0.8\textwidth,trim=4 4 4 4,clip]{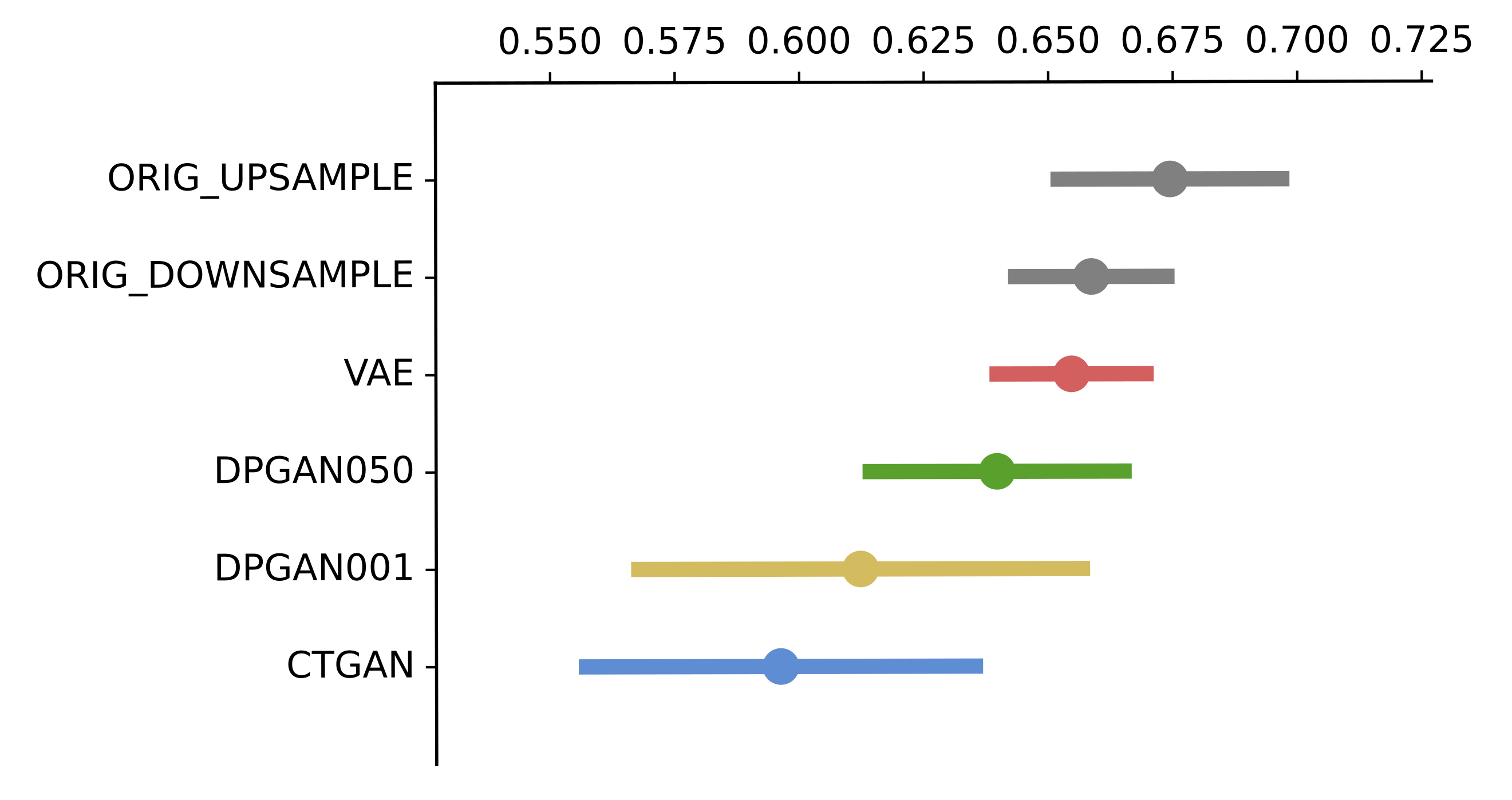}
         \caption{AUC-ROC}
         \label{fig:augauc}
     \end{subfigure}
     \hfill
     \begin{subfigure}[b]{0.5\textwidth}
         \raggedright
         \includegraphics[width=0.8\textwidth,trim=4 4 4 4,clip]{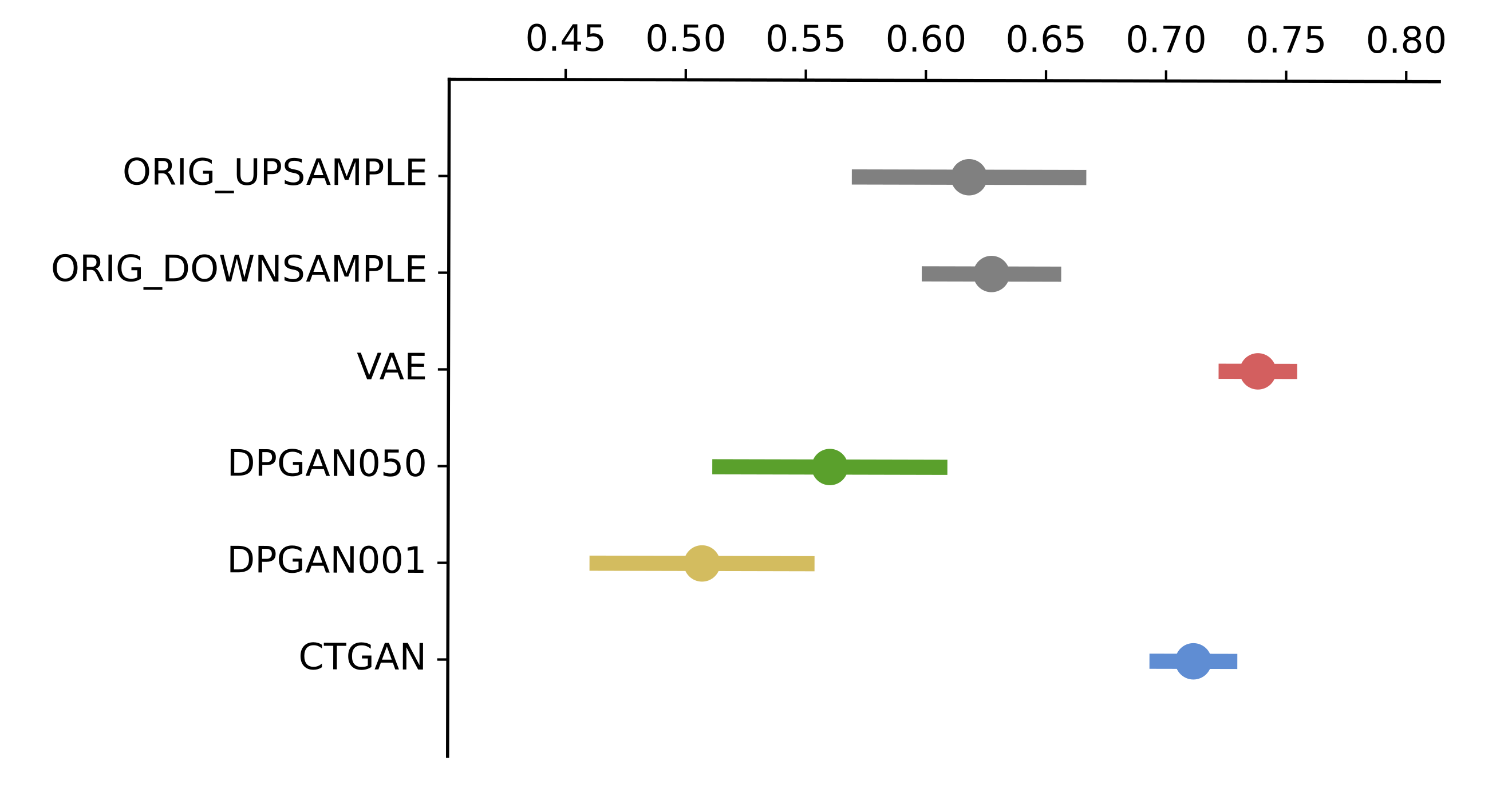}
         \caption{Accuracy}
         \label{fig:augacc}
     \end{subfigure}
     \hfill
     \hfill
     \begin{subfigure}[b]{0.5\textwidth}
         \raggedleft
         \includegraphics[width=0.8\textwidth,trim=4 4 4 4,clip]{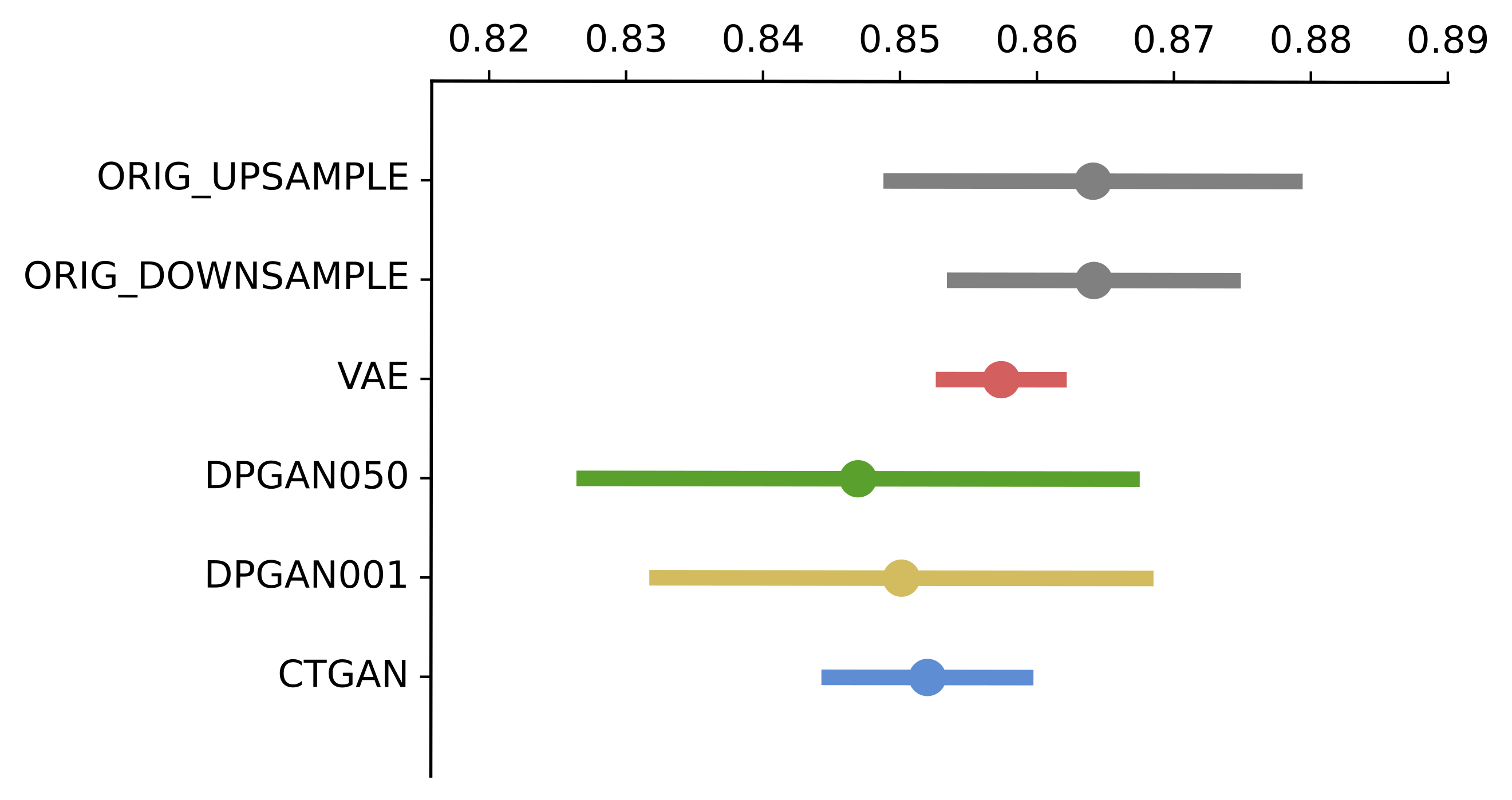}
         \caption{Precision}
         \label{fig:aucprec}
     \end{subfigure}
     \begin{subfigure}[b]{0.5\textwidth}
         \raggedright
         \includegraphics[width=0.8\textwidth,trim=4 4 4 4,clip]{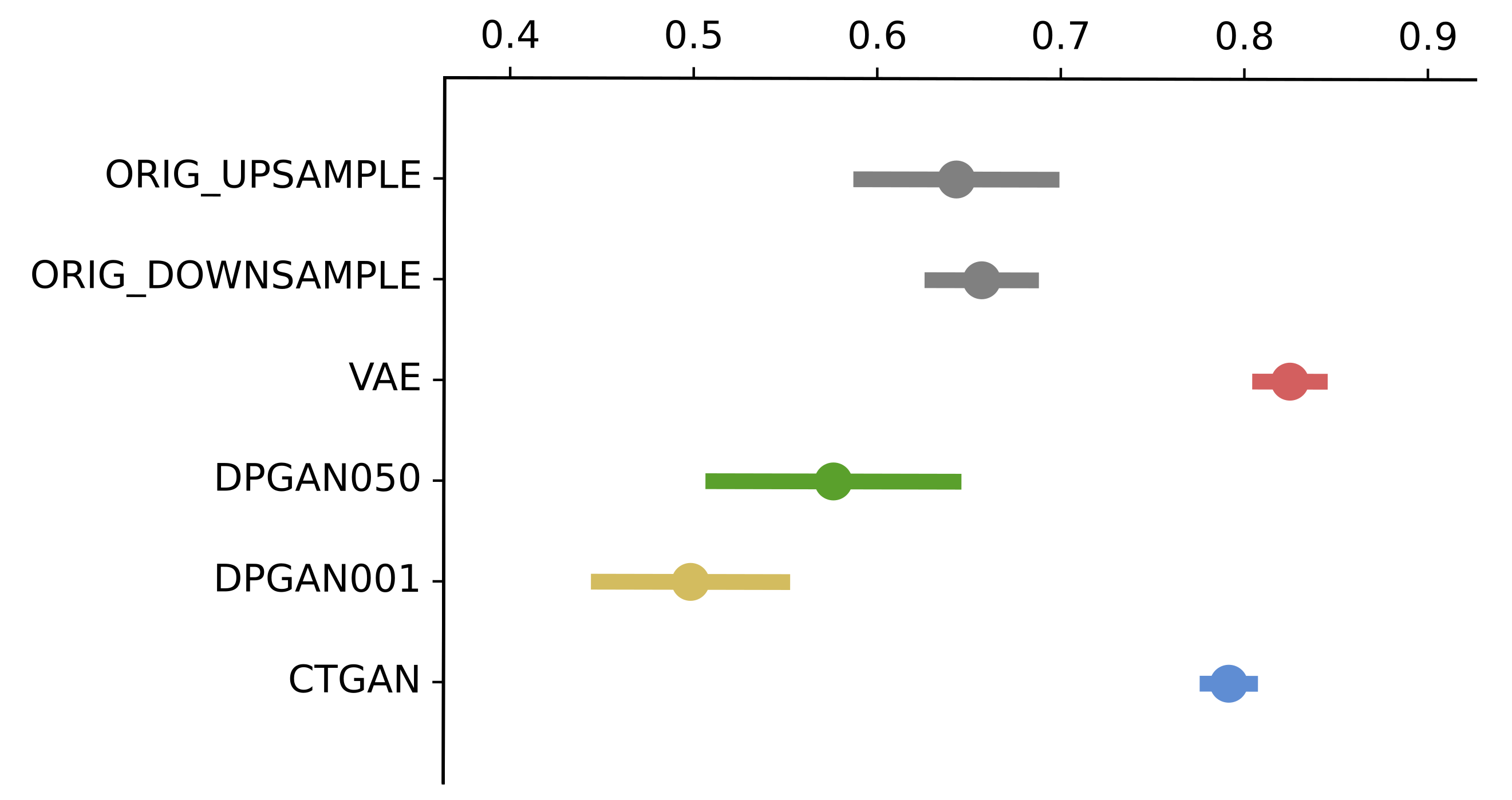}
         \caption{Recall}
         \label{fig:augrec}
     \end{subfigure}
    \caption{Performance of original data augmented with synthetic samples. For each fold, synthetic samples are randomly drawn to augment the small class. Where there are not enough of the small class, the large class is randomly downsampled achieveing class balance. Gaussian Naive Bayes, Random Forest, Neural Network, SVM and Logistic Regression classifiers are each trained to predict the hold out test set outcome and results which maximise performance metrics are reported, averaged over each fold and repeated 10 times.The average number of samples for each model and class is $2128$, $419$, $1029$, $1908$, $1974$ and $2128$ for Upsampling, Downsampling, VAE, DPGAN050, DPGAN001, CTGAN respectively.}
    \label{fig:augmenting}
\end{figure*}

\section{Results}\label{results}
\paragraph{Uniqueness. }Figure \ref{fig1:privacy} shows the average per fold proportion of duplicated data from generation of $100,000$ samples. Of $100,000$ generated samples, VAE has the highest duplication rate of the original training data ($67\%$), followed by the CTGAN ($34\%$), DPGAN001 ($32\%$) and DPGAN050 ($18\%$), as shown in dark grey. Although DPGAN050 has the lowest duplication rate of the original training data, its novel synthesised samples are largely duplicated (approx $71$k duplicates shown in light purple), compared to $26$k, $55$k and $31$k for VAE, DPGAN001 and CTGAN, respectively. CTGAN generates the largest number of unique samples at an average of $33,689$ and VAE has the fewest with $6,647$ (dark purple). Both DPGAN models generate a similar number of unique samples ($12,481$ for DPGAN001 and $11,288$ for DPGAN050). 

\paragraph{Similarity. }Random subsets from each unique synthetic dataset are sampled and similarity with the original data is computed (shown in Figure \ref{fig2:similarity}). Batches are sampled to match the size of the training data while preserving the outcome class ratio. Synthesiser VAE has the highest precision ($0.95\pm0.01$) and density ($0.63\pm0.03$). Given that this model has the largest copying rate of the original data, it follows that its unique data also lies closest to the original data, preserving fidelity. Synthesiser CTGAN has the lowest fidelity (both precision $0.37\pm0.04$ and density $0.13\pm0.02$) with the original training data, however, attains the highest recall ($0.67\pm0.02$). When measuring diversity using coverage, CTGAN displays the largest degredation ($0.31\pm0.02$) placing last with DPGAN050 ($0.31\pm0.02$). This suggests that many of the diverse samples generated by CTGAN are considered outliers (do not match the density of the original distribution). VAE also displays a marked reduction in diversity, compared to both DPGAN models, when measured with recall ($0.58\pm0.03$) versus coverage ($0.43\pm0.01$). DPGAN001 scores consistently higher than DPGAN050 across all similarity metrics ($0.64\pm0.09$ vs $0.54\pm0.05$ for precision, $0.46\pm0.12$ vs $0.31\pm0.02$ for recall, $0.28\pm0.05$ vs $0.22\pm0.02$ for density and $0.44\pm0.10$ vs $0.31\pm0.12$ for coverage).

\paragraph{Classification. } None of the synthetic datasets reach an AUC-ROC that is higher than the original training data. VAE has the highest AUC of all synthetic datasets, which is in line with it ranking the highest across similarity metrics. Model rank across the sum of similarity metrics is VAE, DPGAN001, CTGAN and finally DPGAN050 which does not follow the ranking of AUC-ROC scores. 

\paragraph{Synthetic augmentation to balance classes. } We compared performance measures of a balanced dataset by both upsampling and downsampling, to a balanced dataset augmented with synthetic samples (Figure \ref{fig:augmenting}). The ordering of AUC-ROC scores is preserved from the synthetic data only classification task (Figure \ref{fig:augauc} versus Figure \ref{fig3:synth_auc}) with a smaller difference to the original baseline performance. In contrast, synthetically augmented data from VAE and CTGAN outperform the baseline models when measured using accuracy (Figure \ref{fig:augacc}). This increased performance is largely driven by recall (true positive rate) (Figure \ref{fig:augrec}).

\begin{figure*}
     \begin{subfigure}[b]{0.19\textwidth}
         \centering
         \includegraphics[width=\textwidth]{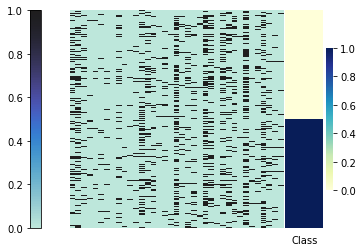}
         \caption{Oversampled}
         \label{fig1:oversampled_heatmap}
     \end{subfigure}
     \hfill
     \begin{subfigure}[b]{0.19\textwidth}
         \centering
         \includegraphics[width=\textwidth]{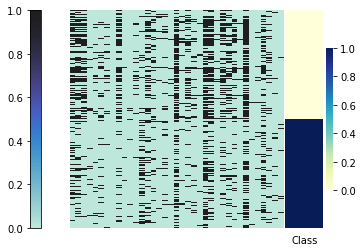} 
         \caption{VAE}
         \label{fig2:vae_heatmap}
     \end{subfigure}
     \begin{subfigure}[b]{0.19\textwidth}
         \centering
         \includegraphics[width=\textwidth]{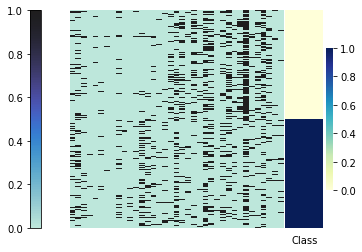}
         \caption{DPGAN050}
         \label{fig3:dpgan050_heatmap}
     \end{subfigure}
     \begin{subfigure}[b]{0.19\textwidth}
         \centering
         \includegraphics[width=\textwidth]{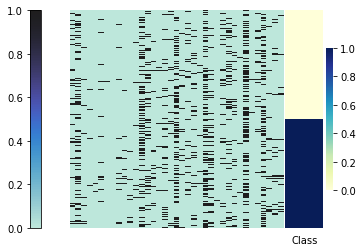}
         \caption{DPGAN001}
         \label{fig3:dpgan001_heatmap}
     \end{subfigure}
     \begin{subfigure}[b]{0.19\textwidth}
         \centering
         \includegraphics[width=\textwidth]{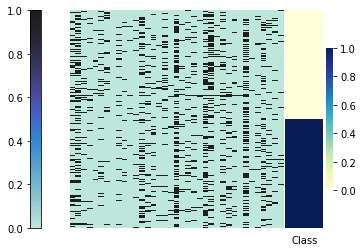}
         \caption{CTGAN}
         \label{fig3:ctgan_heatmap}
     \end{subfigure}
     \hfill
     \begin{subfigure}[b]{0.19\textwidth}
         \centering
         \includegraphics[width=0.9\textwidth]{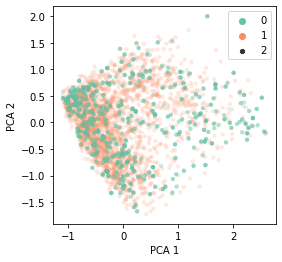}
         \caption{Oversampled}
         \label{fig1:oversampled_pca}
     \end{subfigure}
     \hfill
     \begin{subfigure}[b]{0.19\textwidth}
         \centering
         \includegraphics[width=0.9\textwidth]{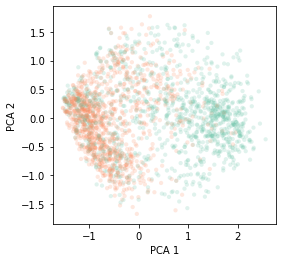} 
         \caption{VAE}
         \label{fig2:vae_pca}
     \end{subfigure}
     \begin{subfigure}[b]{0.19\textwidth}
         \centering
         \includegraphics[width=0.9\textwidth]{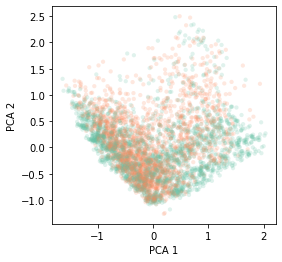}
         \caption{DPGAN050}
         \label{fig3:dpgan050_pca}
     \end{subfigure}
     \begin{subfigure}[b]{0.19\textwidth}
         \centering
         \includegraphics[width=0.9\textwidth]{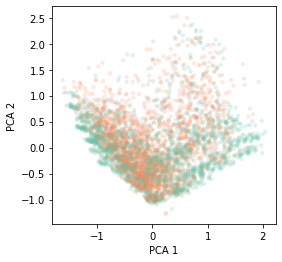}
         \caption{DPGAN001}
         \label{fig3:dpgan001_pca}
     \end{subfigure}
     \begin{subfigure}[b]{0.19\textwidth}
         \centering
         \includegraphics[width=0.9\textwidth]{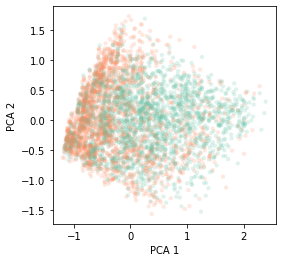}
         \caption{CTGAN}
         \label{fig3:ctgan_pca}
     \end{subfigure}
    \caption{Top row shows heatmap of binary feature matrices for each synthesised dataset. Class 0 is top half of matrix and Class 1 is bottom half. Note that bottom half is from original data while top half is augmented with synthetic data. This balanced dataset serves as input for the classification algorithms as per \nameref{results} section. The bottom row shows 2D PCA embeddings of the same data.}
    \label{fig:sampling_bias}
\end{figure*}

\section{Discussion}\label{discussion} 

Synthetic data augmented with the original data from models VAE and CTGAN outperform the original data accuracy when tested on the hold out set. This is driven by recall, indicating that these models are better able to identify the positive class with little degradation to precision. On average VAE and CTGAN make $511$ and $494$ predictions for the positive class while there are on average $532$ out of $637$ positive samples in the test set. This compared to DPGAN001 and DPGAN050 which make $300$ and $341$ predictions for the positive class. These classification models have learnt a set of features which predict each class with greater accuracy than both DPGAN models and simply resampling (baseline). We observe that CTGAN has the greatest number of unique samples (Figure \ref{fig1:privacy}) and has the greatest diversity when measured using recall (Figure \ref{fig2:similarity}). This diversity of CTGAN lends itself to extracting generalisable features which predict each class. While VAE produces far fewer unique samples, it retains diversity in its samples. In order to better understand the nature of the signal identified by these models, we have plotted the heatmap of the binary feature matrix for one fold (see Figure \ref{fig:sampling_bias}). In the top row of figures, the lower half of the heatmap is coming from the original distribution, while the upper half is synthetic data augmented with the original data for class balance. Here we observe that synthesiser VAE has largely exaggerated the signal for features in the class 0. This has resulted in greater separability in the input space, as shown by the 2D PCA reduction in Figure \ref{fig2:vae_pca}. Reduced separability is displayed by both DPGAN models (Figure \ref{fig3:dpgan001_pca} and \ref{fig3:dpgan050_pca}) with DPGAN001 reporting a lower accuracy. The heatmaps go some way in explaining this since there does not appear to be a marked difference in the upper and lower halves of the DPGAN001 heatmap (Figure \ref{fig3:dpgan001_heatmap}). DPGAN050 however, does show some exaggerated features for class 0 (Figure \ref{fig3:dpgan050_heatmap}). Visually, CTGAN appears to have the most similar heatmap as the original data upsampled, while obtaining greater separability (Figure \ref{fig3:ctgan_pca}). This visual heuristic of similarity appears to be in contrast to the similarity metrics reported in the previous section. This discrepancy could be on account of two things. Since our similarity metrics measure both the positive and negative outcome class in proportions with which they appear in the original training set, the similarity of the larger positive class could have far outweighed that of the smaller negative class. On the other hand, it may be that our similarity metrics are simply not robust at capturing the true similarity of our data. Another method for measuring precision and recall has been proposed in \cite{alaa2021faithful} displaying greater adherence to utility ranking than those measures described in this paper. Unfortunately, we did not obtain code for their metrics, called $\alpha$-precision and $\beta$-precision. These metrics essentially limit the support of each distribution by defining an $\alpha$ or $\beta$ radius on embedded hypersphere of data.

\section{Conclusion}
We examine the utility, similarity and uniqueness of four synthetic data generators and augment the original data with synthetic data to enhance predictability in patient outcome. We observed increased accuracy in performance for both VAE and CTGAN synthetic data generators. Similarity metrics appear to go a little way in explaining the performance of synthetic data generators. Furthermore, while the amplification of a signal in the synthetic dataset may do poorly to preserve faithfulness of the original data, it can provide greater separability hence predictive performance. Considering also the uniqueness of each synthetic data generator, synthesiser CTGAN offers both high predictive performance and uniqueness of samples. This is beneficial for considering stricter conditions on privacy. These trade-offs are problem specific and conclusions are to be arrived at based on clinical relevance. For example, do exaggerated signals corroborate clinical evidence for co-occurence of diseases? Given the caveats of our dataset, we have not shown the clinical relevance of individual features. We have compared the efficacy of generative models to augment imbalanced data for increased predictive performance.

\clearpage
\appendix
\begin{table*}
\centering
\section{Appendix A: Cystic Fibrosis Patient Descriptives}
\label{baseline}
\caption{Baseline characteristics of patients with Cystic Fibrosis in the IBM Explorys EHR dataset. $\Delta$ columns show IBM statistics less statistics from UK CF Registry from \citep{alaa2018autoprognosis} and the US CF Registry \citep{USCF}. Variables with strikethrough are not included in analysis either because there is a high proportion of missingness or because the encoding is uncertain (see footnotes). Codes used are available on request.}
\tiny
\hspace*{0.4cm}
\begin{tabular}{|m{2.2cm}|m{1.3cm}|m{1.3cm}|m{0.8cm}|m{0.8cm}|m{2.2cm}|m{1.3cm}|m{1.3cm}|m{0.8cm}|m{0.8cm}|}
\hline
\rowcolor[rgb]{0.753,0.753,0.753} \textbf{Variable}  & \textbf{Alive \& no LT\linebreak n = 8781 (\%)}                                             & \textbf{Death/LT \linebreak n = 1293 (\%)} & \textbf{UKCF$\Delta$} & \textbf{USCF$\Delta$} & \textbf{Variable}                                            & \textbf{Alive \& no LT\linebreak n = 8781 (\%)} & \textbf{Death/LT \linebreak n = 1293 (\%)} & \textbf{UKCF$\Delta$} & \textbf{USCF$\Delta$}  \\ 
\hhline{----------}
\textbf{Gender(\%male)} & 4,925 (56.1) & 622 (48.1\%)  &  &  & \textit{Pancreatic } &  &  &   &  \\
\textbf{\sout{Age (years)§}} & \multicolumn{2}{c|}{55.6\% missing} &  &  & 
\hspace{1mm} Cirrhosis & 125 (1.4) & 32 (2.5) & (1.2) & (1.6) \\
\textbf{\sout{Height (cm)§}} & \multicolumn{2}{c|}{81.0\% missing} &  &  & 
\hspace{1mm} Liver Disease & 309 (3.5) & 68 (5.3) & (12.5)  & (-0.1) \\
\textbf{\sout{Weight (kg)§}} & \multicolumn{2}{c|}{80.0\% missing} &  &  & 
\hspace{1mm} \sout{Pancreatitis}$^c$ & 11 (0.1)  & 2 (0.2)  & (1.3)  & (0.9) \\
\textbf{\sout{BMI (kg/m2)§}} & \multicolumn{2}{c|}{81.0\% missing}  &  &  & 
\hspace{1mm} \sout{Liver Enzymes}$^d$  & 5 (0.05)  & 0 (0) & (15.1) &  \\
\textbf{Genotype} &  &   &   &   & 
\hspace{1mm} Gall Bladder & 96 (1.1)  & 17 (1.3)  & (-0.6) &   \\
\hspace{1mm} Homozygous  & \multicolumn{2}{c|}{Not EHR} 
&  &  & 
\hspace{1mm} GI Bleed (variceal) & \multicolumn{2}{c|}{Not found} &   & \\
\hspace{1mm} Heterozygous & \multicolumn{2}{c|}{Not EHR}  &   &  & \textit{Gastrointestinal}  &  &   &   &   \\
\hspace{1mm} $\Delta$F508  & \multicolumn{2}{c|}{Not EHR}  &   &  & \hspace{1mm} GERD  & 1675 (19.1)  & 269 (20.8)   & (1.6)  & (17.4) \\
\hspace{1mm} G551D  & \multicolumn{2}{c|}{Not EHR}   &    &    & \hspace{1mm} GIB (no variceal)   & 168 (2.0)  & 43 (3.3)  & (-2.0) &   \\
\hspace{1mm} Class I & \multicolumn{2}{c|}{Not EHR}  &  &   & 
\hspace{1mm} Intestinal Obstruction & 358 (4.1)  & 125 (9.7)  & (3.6)  &   \\
\hspace{1mm} Class II  & \multicolumn{2}{c|}{Not EHR}  &   &    & \textit{Musculoskeletal } &   &    &  & \\
\hspace{1mm} Class III  & \multicolumn{2}{c|}{Not EHR}  &  &  & \hspace{1mm} Arthropathy  & 440 (5.0)  & 62 (4.8) & (4.6) & (-1.8) \\
\hspace{1mm} Class IV & \multicolumn{2}{c|}{Not EHR} &  &   & \hspace{1mm} Bone Fracture  & 131 (1.5) & 24 (1.9)  & (-0.4) & (-1.3)  \\
\hspace{1mm} Class V & \multicolumn{2}{c|}{Not EHR}  &   &  & \hspace{1mm} Osteopenia & 4 (0.05)  & 1 (0.1)  & (20.6)  & (9.9) \\
\hspace{1mm} Class VI & \multicolumn{2}{c|}{Not EHR}   &  &  & \textit{Other}  &  &   &   &  \\
\textbf{Spirometry §}  &  &   &   &   & 
\hspace{1mm} Cancer & 17 (0.2) & 6 (0.5)  & (0.1) & (-0.1)  \\
\hspace{1mm} \sout{FEV1 (L)} & \multicolumn{2}{c|}{99.3\% missing}  &  &  & 
\hspace{1mm} Diabetes & 1592 (18.1)  & 245 (18.9)  &  (9.0)  &   \\
\hspace{1mm} \sout{FEV1\%}  & \multicolumn{2}{c|}{97.6\% missing}  &   &   & 
\hspace{1mm} CFRD & 789 (9.0) & 153 (11.8)  & (23.2)  & (9.3) \\
\hspace{1mm} \sout{Best FEV1 (L)} & \multicolumn{2}{c|}{99.3\% missing} &   &  & 
\hspace{1mm} Pulmonary Abscess & 32 (0.4) & 15 (1.2)  & (-0.4)  &  \\
\hspace{1mm} \sout{Best FEV1\%}  & \multicolumn{2}{c|}{98.7\% missing}  &  &   & 
\hspace{1mm} Chr. Pseudomonas & 549 (6.3)  & 148 (11.4)  &  (49.5) &  \\
\hspace{1mm} \sout{FEV1\% (2017)} & \multicolumn{2}{c|}{99.5\% missing}  &   &    & \hspace{1mm} Osteoporosis & 497 (5.7) & 129 (10.0)  & (3.4) & (-2.1) \\
\hspace{1mm} \sout{FEV1\% (2016)} & \multicolumn{2}{c|}{99.5\% missing}  &   &  & 
\hspace{1mm} AICU & \multicolumn{2}{c|}{Not found}   &     &\\
\hspace{1mm} \sout{FEV1\% (2015)}  & \multicolumn{2}{c|}{99.5\% missing}  &  &  & 
\hspace{1mm} Kidney Stones & 439 (5.0)  & 79 (6.0)  & (-3.6)  & (-4.4)  \\
\hspace{1mm} \sout{FEV1\% (2014)} & \multicolumn{2}{c|}{99.5\% missing}  &   &   & 
\hspace{1mm} Cough Fracture  & 53 (0.6) & 10 (5.5)  & (-0.5)  &  \\
\textbf{Lung Infections}  &  &   &   &  & 
\hspace{1mm} Hypertension  & 1784 (20.3)  & 285 (0.22) & (-14.9) & (-14.6) \\
\hspace{1mm} \sout{B. Cepacia}$^a$ & 7 (0.01)  & 2 (0.1) & (5.2)  & (1.6) & \hspace{1mm} A.Mycobacteria  & 132 (1.5) & 18 (1.4)  & (2.0) & (8.5) \\
\hspace{1mm} P. Aeruginosa & 686 (7.8) & 172 (13.3) & (52.8) & (23.7) & \hspace{1mm} Hearing Loss & 255 (2.9) & 59 (4.6) & (-0.4)  & (-0.5) \\
\hspace{1mm} MRSA & 831 (9.5) & 172 (13.3) & (-5.6)   &  (9.7) & 
\hspace{1mm} Depression  & 1203 (13.7) & 161 (12.5) & (-5.8) & (3.4) \\
\hspace{1mm} Aspergillus & 132 (1.5) & 62 (4.8)  & (10.5)  &  & \textbf{Inhaled Antibiotics} &   \multicolumn{2}{c|}{Not found}  &  & \\
\hspace{1mm} NTM & 80 (1.0) & 6 (0.5)  & (4.2)  & (9.0)  & 
\textbf{Muco-active Therapy}  &  &  &   &   \\
\hspace{1mm} H. Influenza & 72 (0.8) & 25 (1.9) & (4.1) & (9.1) & 
\hspace{1mm} DNase  & 252 (2.9) & 17 (1.3)  &  (55.2) & (88.8)  \\
\hspace{1mm} E. Coli & 203 (2.3) & 50 (3.9) & (-2.0) &  & 
\hspace{1mm} Hypertonic Saline & 1038 (11.8) & 209 (16.1) & (11.6)  & (61.3)\\
\hspace{1mm} K. Pneumoniae & 72 (0.8) & 26 (2.0)  & (-0.6)  &  & \textbf{\sout{Promixin}}$^e$ & 9 (0.1) & 2 (0.2) & (20.5) & \\
\hspace{1mm} Gram-negative & 0 & 0 & (0.5) & & \textbf{Tobramycin} & 365 (4.2)  & 70 (5.4) & (-0.9) & (61.0) \\
\hspace{1mm} ALCA & 37 (0.4) & 4 (0.3) &  (2.3) &  & \textbf{iBuprofen} & 433 (5.0) & 32 (2.5) & (-4.5) &  (-3.8) \\
\hspace{1mm} Staph. Aureus & 323 (3.7) & 89 (6.9)  & (19.9)   &  (52.9)  & \textbf{Oral Corticosteroids}   & \multicolumn{2}{c|}{Not found}   &   &  \\
\hspace{1mm} \sout{Xanthomonas}$^b$ & 60 (0.7) & 9 (0.7) & (3.2) &  & 
\textbf{IV Antibiotics} & \multicolumn{2}{c|}{Not found}   &  & \\
\hspace{1mm} B. Multivorans   & \multicolumn{2}{c|}{This is a UK concept}  & &   & \textbf{IV Antibiotic Courses}  &  &   & &  \\
\hspace{1mm} B. Cenocepacia   & \multicolumn{2}{c|}{This is a UK concept}  &  &  & 
\hspace{1mm} Days at Home  & \multicolumn{2}{c|}{Not found} &    & \\
\hspace{1mm} Pandoravirus  & \multicolumn{2}{c|}{Not found}   &  & & 
\hspace{1mm} Days at Hospital & \multicolumn{2}{c|}{Not found} &   & \\
\textbf{Comorbidities}  & &   &  &   & \textbf{Non-IV \linebreak Hospitalization} & \multicolumn{2}{c|}{Not found}  &  & \\
\textit{Respiratory} &  &   &   &   & \textbf{Non-IV Ventilation} & \multicolumn{2}{c|}{Not found}   &   &   \\
\hspace{1mm} ABPA & 138 (1.6)  & 17 (1.3)  & (10.8)  & (3.4)    & \textbf{Oxygen Therapy} & \multicolumn{2}{c|}{Not found}  &  &   \\
\hspace{1mm} Nasal Polyps  & 269 (3.1)  & 35 (2.7)   &  (0.0)  & (6.8) & \hspace{1mm} Continuous  & \multicolumn{2}{c|}{Not found}   &   &   \\
\hspace{1mm} Asthma  & 1604 (18.3) & 151 (11.7)  & (-2.0)  & (13.4)  & \hspace{1mm} Nocturnal  & \multicolumn{2}{c|}{Not found}   &    &    \\
\hspace{1mm} Sinus Disease    & 750 (8.5) & 110 (8.5)  &  (4.5)  & (1.4) & \hspace{1mm} Exacerbation  & \multicolumn{2}{c|}{Not found}   &  &   \\
\hspace{1mm} Hemoptysis & 398 (4.5)  & 85 (6.6) & (-3.2) & (-2.8)   & 
\hspace{1mm} Pro re nata & \multicolumn{2}{c|}{Not found}   &   &    
\\
\hline
\end{tabular}
\label{cf_diff}\footnotesize{
$^a$ LOINC code (44800-1) refers to genus Burkholderia and does not specify Cepacia.
$^b$ Includes Stenotrophomonas maltophilia and not exhaustive of Xanthomonas species.
$^c$ 
$^d$ 
$^e$ Finding not diagnosis.
}\\
\end{table*}
\clearpage

\section{Appendix B: Experimental Details}
\label{experimental_details}

\paragraph{Model Architectures. } 
We implement a differentially private GAN using Opacus pytorch library \cite{opacus}. This adds gaussian noise directly to the parameter gradients as detailed in \cite{xie2018differentially}. Our generator is made up of 4 blocks followed by a fully connected layer and tanh activation function (as shown in Figure \ref{subfig:dpgan_gen}, discriminator architecture shown in Figure \ref{subfig:dpgan_dis}). We train the model with a binary cross entropy loss.

\definecolor{fc}{HTML}{1E90FF}
\definecolor{h}{HTML}{228B22}
\definecolor{bias}{HTML}{87CEFA}
\definecolor{noise}{HTML}{8B008B}
\definecolor{conv}{HTML}{FFA500}
\definecolor{pool}{HTML}{B22222}
\definecolor{up}{HTML}{B22222}
\definecolor{view}{HTML}{FFFFFF}
\definecolor{bn}{HTML}{FFD700}
\tikzset{fc/.style={black,draw=black,fill=fc,rectangle,minimum height=1cm}}
\tikzset{h/.style={black,draw=black,fill=h,rectangle,minimum height=1cm}}
\tikzset{bias/.style={black,draw=black,fill=bias,rectangle,minimum height=1cm}}
\tikzset{noise/.style={black,draw=black,fill=noise,rectangle,minimum height=1cm}}
\tikzset{conv/.style={black,draw=black,fill=conv,rectangle,minimum height=1cm}}
\tikzset{pool/.style={black,draw=black,fill=pool,rectangle,minimum height=1cm}}
\tikzset{up/.style={black,draw=black,fill=up,rectangle,minimum height=1cm}}
\tikzset{view/.style={black,draw=black,fill=view,rectangle,minimum height=1cm}}
\tikzset{bn/.style={black,draw=black,fill=bn,rectangle,minimum height=1cm}}

\begin{figure}[h]
  \centering
  \begin{tikzpicture}
    \node (n) at (.25,0) {\small$z_{\mathcal{N}(0,1)}$};
    \node[bias,rotate=90,minimum width=2cm] (b1) at (1.5,0) {\small$\text{B1}_{z, 128}$};
    \node[bias,rotate=90,minimum width=2cm] (b2) at (2.75,0) {\small$\text{B2}_{128,256}$};
    \node[bias,rotate=90,minimum width=2cm] (b3) at (4,0) {\small$\text{B3}_{256,512}$};
    \node[bias,rotate=90,minimum width=2cm] (b4) at (5.25,0) {\small$\text{B4}_{512, 512}$};
    \node[fc,rotate=90,minimum width=2cm] (d) at (6.5,0) {\small$\text{FC}_{512, 41}$};
    \node[h,rotate=90,minimum width=2cm] (a) at (7.75,0) {\small$\text{tanh}$};
    \node (x) at (9,0) {\small$\tilde{x}$};

    \node[bias,minimum width=5cm,minimum height=2.25cm] (block) at (5.85,-2.5) {\small};
    \node[bias,rotate=90,minimum width=2cm] (block_) at (1.5,-2.5) {\small$\text{Block}$};
    \node[fc,rotate=90,minimum width=2cm] (Ba) at (4,-2.5) {\small$\text{FC}_{in, out}$};
    \node[bn,rotate=90,minimum width=2cm] (Bb) at (5.25,-2.5) {\small$\text{dropout}$};
    \node[conv,rotate=90,minimum width=2cm] (Bc) at (6.5,-2.5) {\small$\text{batch norm}$};
    \node[h,rotate=90,minimum width=2cm] (Bd) at (7.75,-2.5) {\small$\text{ReLu}$};
  
    \draw[->] (n) -- (b1);
    \draw[->] (b1) -- (b2);
    \draw[->] (b2) -- (b3);
    \draw[->] (b3) -- (b4);
    \draw[->] (b4) -- (d);
    \draw[->] (d) -- (a);
    \draw[->] (a) -- (x);

    \draw[-] (block) -- (block_);
    \draw[->] (Ba) -- (Bb);
    \draw[->] (Bb) -- (Bc);
    \draw[->] (Bc) -- (Bd);
    
    \draw[-] (block) -- (block_);
  \end{tikzpicture}
  \vskip 2px
  \caption{Generator of DPGAN synthesiser takes a $z$-dimensional noise vector as input.}
  \label{subfig:dpgan_gen}
\end{figure}
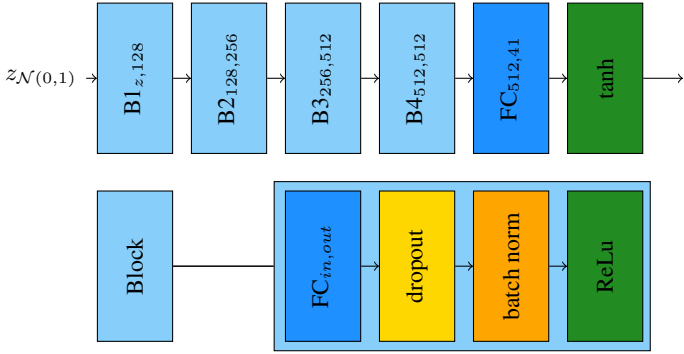

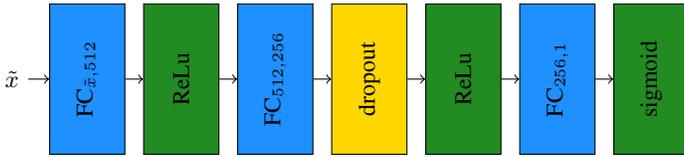
\begin{figure}[h]
  \centering
  \begin{tikzpicture}
    \node (n) at (0,0) {\small$\tilde{x}$};
    \node[fc,rotate=90,minimum width=2cm] (b1) at (1,0) {\small$\text{FC}_{\tilde{x}, 512}$};
    \node[h,rotate=90,minimum width=2cm] (b2) at (2.25,0) {\small$\text{ReLu}$};
    \node[fc,rotate=90,minimum width=2cm] (b3) at (3.5,0) {\small$\text{FC}_{512,256}$};
    \node[bn,rotate=90,minimum width=2cm] (b4) at (4.75,0) {\small$\text{dropout}$};
    \node[h,rotate=90,minimum width=2cm] (d) at (6.0,0) {\small$\text{ReLu}$};
    \node[fc,rotate=90,minimum width=2cm] (a) at (7.25,0) {\small$\text{FC}_{256,1}$};
    \node[h,rotate=90,minimum width=2cm] (s) at (8.5,0) {\small$\text{sigmoid}$};
  
    \draw[->] (n) -- (b1);
    \draw[->] (b1) -- (b2);
    \draw[->] (b2) -- (b3);
    \draw[->] (b3) -- (b4);
    \draw[->] (b4) -- (d);
    \draw[->] (d) -- (a);
    \draw[->] (a) -- (s);
    
  \end{tikzpicture}
  \vskip 2px
  \caption{Discriminator architecture of DPGAN synthesiser.}
  \label{subfig:dpgan_dis}
\end{figure}
\begin{figure}[h!]
  \centering
  \begin{tikzpicture}
    \node (x) at (1.25,0) {\small$x$};
    \node[fc,rotate=90,minimum width=2cm] (fc1) at (2.5,0) {\small$\text{FC}_{41,h}$};
    \node[h,rotate=90,minimum width=2cm] (r1) at (3.75,0) {\small$\text{ReLu}$};
    \node[bn,rotate=90,minimum width=2cm] (d1) at (5,0) {\small$\text{dropout}$};
    \node[fc,rotate=90,minimum width=2cm] (fc2) at (6.25,0) {\small$\text{FC}_{h, h}$};
    \node[h,rotate=90,minimum width=2cm] (r2) at (7.5,0) {\small$\text{ReLu}$};
    \node[fc,rotate=90,minimum width=2cm] (fc3) at (8.75,0) {\small$\text{fc}_{h, z}$};
    \node (z) at (9.75,-2.5) {\small$z$};

    \node[h,rotate=90,minimum width=2cm] (s) at (1.25,-2.5) {\small$\text{sigmoid}$};
    \node[fc,rotate=90,minimum width=2cm] (fc6) at (2.5,-2.5) {\small$\text{FC}_{h, 41}$};
    \node[h,rotate=90,minimum width=2cm] (r4) at (3.75,-2.5) {\small$\text{ReLu}$};
    \node[fc,rotate=90,minimum width=2cm] (fc5) at (5,-2.5) {\small$\text{FC}_{h, h}$};
    \node[bn,rotate=90,minimum width=2cm] (d2) at (6.25,-2.5) {\small$\text{dropout}$};
    \node[h,rotate=90,minimum width=2cm] (r3) at (7.5,-2.5) {\small$\text{ReLu}$};
    \node[fc,rotate=90,minimum width=2cm] (fc4) at (8.75,-2.5) {\small$\text{FC}_{z, h}$};
  
    \node (rx) at (1.25,-1) {\small$\tilde{x}$};
  
    \draw[->] (x) -- (fc1);
    \draw[->] (fc1) -- (r1);
    \draw[->] (r1) -- (d1);
    \draw[->] (d1) -- (fc2);
    \draw[->] (fc2) -- (r2);
    \draw[->] (r2) -- (fc3);
    \draw[-] (fc3) -- (9.75,0);
    \draw[->] (9.75,0) -- (z);

    \draw[->] (z) -- (fc4);
    \draw[->] (fc4) -- (r3);
    \draw[->] (r3) -- (d2);
    \draw[->] (d2) -- (fc5);
    \draw[->] (fc5) -- (r4);
    \draw[->] (r4) -- (fc6);
    \draw[->] (fc6) -- (s);
    \draw[->] (s) -- (rx);

  \end{tikzpicture}
  \vskip 2px
  \caption{Variational Autoencoder Synthesiser architecture.}
  \label{subfig:deep-learning-auto-encoder}
\end{figure}
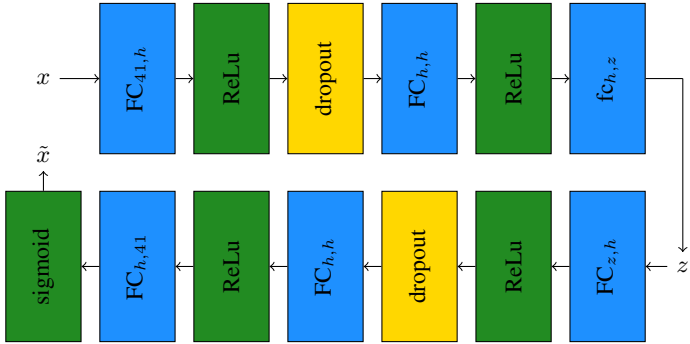
We implement CTGAN by adopting the model architecture provided in \cite{xu2019modeling} and the corresponding code repo. Our variational autoencoder model is shown in Figure \ref{subfig:deep-learning-auto-encoder} trained with a binary cross entropy reconstruction loss and Kullback Liebler (KL) regularisation. 

\paragraph{Hyperparameter Optimisation. } We perform hyperparameter optimisation (HO) for the models using grid search with optimisation objective to maximise the sum of precision, recall, density and coverage. We performed optimisation a stratified (by outcome class) sample of $80\%$ of the data. The final parameters for DPGAN001 are batch size of $150$, latent dimensions $32$, sigma $0.01$, gradient clip $0.1$, lr $0.002$ with decay $0.3$ and $0.999$. For DPGAN050 we used batch size of $50$, latent dimension $10$, sigma $0.5$, gradient clip $0.05$, lr $0.002$ with decay $0.8$ and $0.8$. For the VAE we have batch size $150$, latent dimension $16$, hidden dimension $500$ and lr $0.001$. CTGAN has batch size $300$, generator latent dimension $50$, discriminator latent dimension $512$ and embedding dimension $z=32$. For each generator update we perform $7$ discriminator updates and group $20$ samples together when applying discriminator. We use discriminator learning rate of $0.0002$ and decay $1e-6$, and generator learning rate of $0.002$ and decay of $1e-7$.

\begin{figure}[h!]
    \centering
    \includegraphics[scale=0.7]{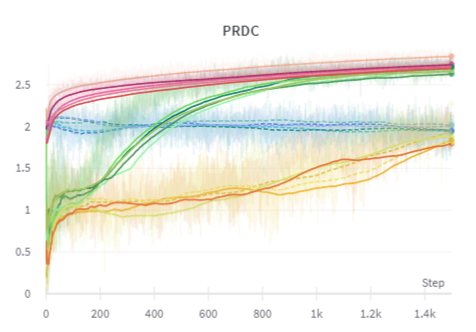}
    \caption{PRDC metrics per training epoch for each model and fold. VAE is shown in pink, DPGAN001 in green, DPGAN050 in blue and CTGAN in orange.}
    \label{fig:prdc}
\end{figure}

In Table \ref{tab:prdc} we observe that the maximum achieved from HO is very similar as that from each fold for both DPGAN models. VAE obtains slightly higher than HO, whereas CTGAN is considerably lower. That CTGAN has not obtained the maximum PRDC from HO could be due to loss of generalisability across folds since unstable combinations of parameters achieve maximum. We have not investigated this further since the training curved across folds appear stable (see Figure \ref{fig:prdc}).

\begin{table}[h!]
    \centering
        \begin{tabular}{lcc} \hline
        Model & PRDC (HO) & PRDC folds\\ \hline
        DPGAN001 & [1.35,2.75] & $2.80\pm0.03$ \\
        DPGAN050 & [0.84,2.04] & $2.00\pm0.20$ \\
        CTGAN & [1.18,2.72] & $2.07\pm 0.04$ \\
        VAE & [2.12,2.63] & $2.78\pm0.05$ \\
        \hline
        \end{tabular}
    \caption{Range of PRDC values from hyperparametisation optimisation (HO). Final mean and std deviation reported in the final column.}
    \label{tab:prdc}
\end{table}

\clearpage

\end{document}